%% file: main.tex
\documentclass[10pt,twocolumn,letterpaper]{article}

\usepackage[pagenumbers]{cvpr}  %

\input{preamble}
\usepackage{tikz}
\usepackage{pifont}
\usepackage{algorithm}
\usepackage{algpseudocode}
\usepackage{multirow}
\usepackage{longtable}
\usepackage{graphicx}

\definecolor{cvprblue}{rgb}{0.21,0.49,0.74}
\usepackage[pagebackref,breaklinks,colorlinks,allcolors=cvprblue]{hyperref}

\DeclareRobustCommand{\cmark}{\ding{51}}
\DeclareRobustCommand{\xmark}{\ding{55}}
\DeclareRobustCommand*\circled[1]{%
  \tikz[baseline=(char.base)]{
    \node[shape=circle,draw,inner sep=0.5pt, line width=0.4pt,
          minimum size=6.5pt] (char) {\fontsize{6.5}{7}\selectfont #1};}}

\newcommand*\fixedcircled[2][1.5em]{%
  \tikz[baseline=(char.base)]{
    \node[shape=circle,draw,inner sep=0pt,minimum size=#1] (char) {\strut #2};}}

\title{ArtiBench and ArtiBrain: Benchmarking Generalizable Vision-Language Articulated Object Manipulation}

\author{
Yuhan Wu$^{1,*}$ \quad
Tiantian Wei$^{2,*}$ \quad
Shuo Wang$^{1}$ \quad
ZhiChao Wang$^{1}$  \\
Yanyong Zhang$^{1}$ \quad
Daniel Cremers$^{2}$ \quad
Yan Xia$^{1,\dagger}$ \\
$^1$University of Science and Technology of China \\
$^2$Technical University of Munich
}

\begin{document}

\twocolumn[{%
\renewcommand\twocolumn[1][]{#1}%

\maketitle

\begin{center}
\centering
\captionsetup{type=figure}
\includegraphics[width=0.95\textwidth]{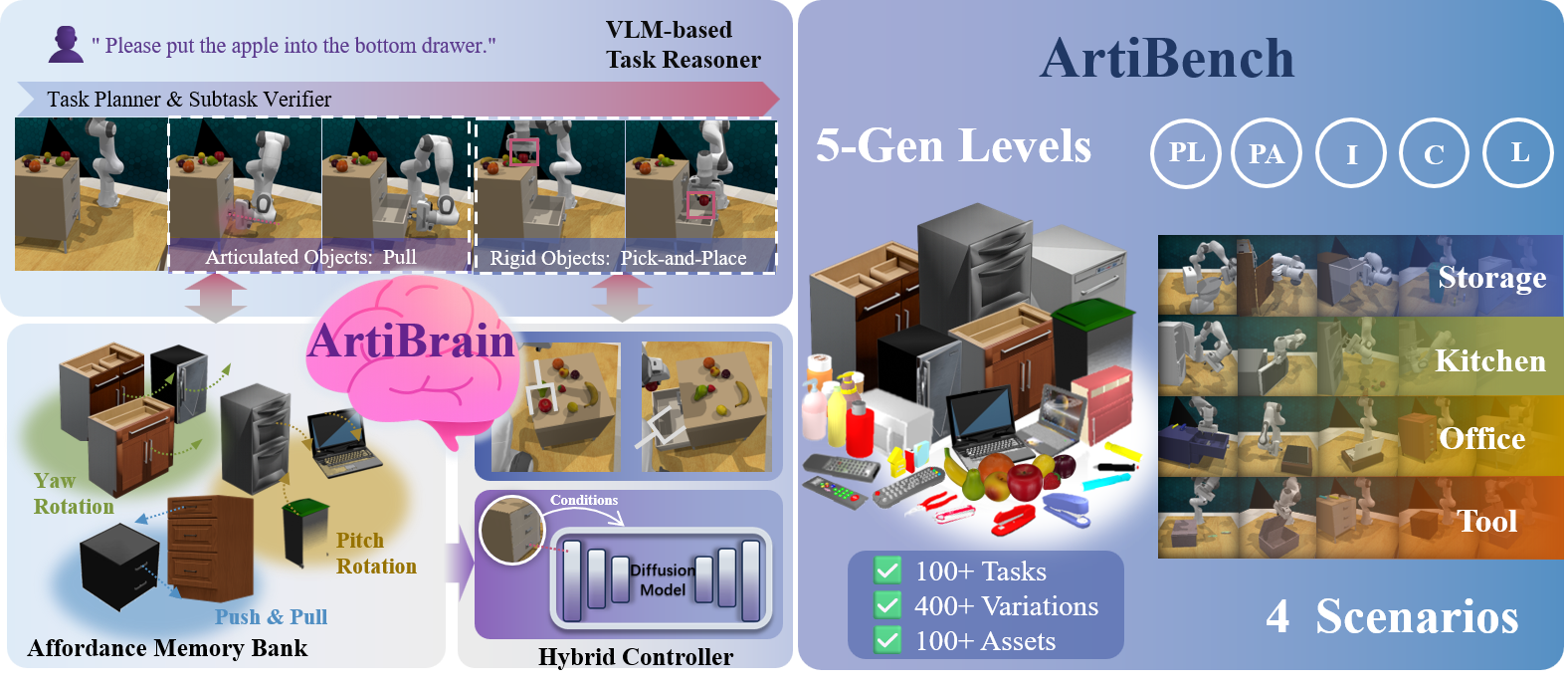}
\captionof{figure}{
Overview of ArtiBrain and ArtiBench.
\textit{(Left)} ArtiBrain performs long-horizon articulated manipulation via hierarchical reasoning and hybrid control. It is a hierarchical, closed-loop framework that integrates three key modules: a VLM-based Task Reasoner, a Hybrid Controller for both rigid and articulated actions, and an Affordance Memory Bank that accumulates verified part-level affordances to enhance transfer across parts and categories. 
\textit{(Right)} ArtiBench provides 100+ articulated tasks and 400+ variations across four household scenarios and five generalization levels, enabling systematic evaluation from part-level variation to long-horizon multi-object manipulation.
}
\label{fig:artibrain-overview}
\end{center}

}]
\renewcommand{\thefootnote}{\fnsymbol{footnote}}
\footnotetext[1]{Equal contributions.}
\footnotetext[2]{Corresponding author.}
\renewcommand{\thefootnote}{\arabic{footnote}}

\input{sec/0_abstract}    
\input{sec/1_intro}

\input{sec/2_related_work}

\input{sec/3_benchmark}

\input{sec/4_method}

\input{sec/5_experiments}

\input{sec/7_conclusion}

{
    \small
    \bibliographystyle{ieeenat_fullname}
    \bibliography{main}
}

\end{document}

%% file: sec/0_abstract.tex
\begin{abstract}
Interactive articulated manipulation requires long-horizon, multi-step interactions with appliances while maintaining physical consistency. Existing vision-language and diffusion-based policies struggle to generalize across parts, instances, and categories.
We first introduce ArtiBench, a five-level benchmark covering kitchen, storage, office, and tool environments. ArtiBench enables structured evaluation from cross-part and cross-instance variation to long-horizon multi-object tasks, revealing the core generalization challenges of articulated object manipulation.
Building on this benchmark, we propose ArtiBrain, a modular framework that unifies high-level reasoning with adaptive low-level control. ArtiBrain uses a VLM-based Task Reasoner (GPT-4.1) to decompose and validate subgoals, and employs a Hybrid Controller that combines geometry-aware keyframe execution with affordance-guided diffusion for precise and interpretable manipulation. An Affordance Memory Bank continually accumulates successful execution episodes and propagates part-level actionable affordances to unseen articulated parts and configurations.
Extensive experiments on ArtiBench show that our ArtiBrain significantly outperforms state-of-the-art multimodal and diffusion-based methods in robustness and generalization. Code and dataset will be released upon acceptance.

\end{abstract}

%% file: sec/1_intro.tex
\vspace{-1.2em}
\section{Introduction}
\label{sec:intro}

Recent advances in robot learning have enabled robots to grasp, place, and rearrange rigid objects in controlled environments~\cite{embodiment2025openx, khazatsky2025droid, jiang2023vima, garcia20253dlotus, goyal2022ifor, guo2024phygrasp, ai2025learningdyn}.
However, real-world tasks often require sequential interactions with articulated objects such as drawers, cabinets, and appliances.
These tasks demand more than goal understanding. They require reasoning about articulation states, contact dynamics, and feasible action sequences.
For instance, the instruction “clean the desk” involves opening a drawer, retrieving an item, and placing it back.
Each step depends on grounding object states and maintaining physical consistency.

Despite progress in robot learning, a gap remains between controlled demonstrations and general embodied intelligence.
To act autonomously in diverse articulated environments, robots must interpret open-ended language instructions, decompose them into sub-tasks, and execute them reliably.
Recent work has improved language-conditioned planning~\cite{huang2023voxposer, liang2022code, zitkovich2023rt2, intelligence2025pi05, duan2024manipulate, li2024manipllm, brohan2023can} and affordance-driven control~\cite{wu2025afforddp, zhao2025anchordp3}.
However, most methods remain limited to rigid-object manipulation or single-step articulated actions~\cite{shridhar2022cliport, jang2022bcz, du2023learning, bousmalis2024robocat, zitkovich2023rt2, chi2023diffusion, ze2024dp3, ke2024diffuseractor, team2024octo, zhao2025anchordp3, belkhale2024rth}.
Existing datasets~\cite{james2019rlbench, mu2021maniskill, gu2023maniskill2,  mees2022calvin, xiang2020sapien, li2024behavior, embodiment2025openx, khazatsky2025droid, walke2023bridgedata} rarely evaluate cross-part, cross-instance, or cross-category generalization.
As a result, there is still no unified framework that connects high-level reasoning with reliable low-level execution in articulated environments.

To address this gap, we introduce ArtiBench, a benchmark for articulated-object manipulation. It contains \textbf{132} articulated scenes and \textbf{449} task variations across \textbf{four} household domains. It defines \textbf{five} generalization levels covering random placement, cross-part, cross-instance, cross-category, and long-horizon composition, enabling systematic evaluation beyond single-step drawer-opening tasks.
Furthermore, we observe that long-horizon articulated tasks couple articulated interactions with rigid-object manipulation, such as picking an item and placing it into a drawer. 
Rigid-motion phases exhibit predictable geometry and can be efficiently handled through structured keyframe or motion-planning strategies, 
whereas articulated interactions demand adaptive, contact-aware control. 
Diffusion-based visuomotor policies excel at capturing contact dynamics but require large-scale expert demonstrations and suffer from sampling latency~\cite{wolf2025diffusionrobotics}. 
In contrast, keyframe-based control offers compact policy representations that generalize well in structured motion regimes~\cite{akgun2012keyframe, yuan2023m2t2}. 

Motivated by these observations, we propose ArtiBrain, a hierarchical policy that unifies high-level reasoning and adaptive low-level control for articulated-object manipulation.
Inspired by the compositional structure of everyday tasks, we design a VLM-based Task Reasoner that parses natural-language instructions and verifies the scene state to produce structured subgoals with explicit success conditions.
To execute these subgoals, we introduce a Hybrid Controller that dynamically switches between two control modes: a geometry-guided keyframe policy (GeoKeyframe) for structured rigid-motion phases, and an affordance-guided diffusion policy for contact-rich articulated interactions (ArtiDiffusion). 
Furthermore, we observe that articulated parts can share transferable affordance priors across categories. 
To leverage this finding, we develop an Affordance Memory Bank that stores and updates part-level priors from successful episodes, enabling robust generalization to unseen objects.

To summarize, the main contributions of this work are:
\begin{itemize}
    \item We address the underexplored problem of long-horizon articulated object manipulation, where robots must jointly reason about part states, contact dynamics, and sequential actions under open-vocabulary instructions.
    
    \item We introduce ArtiBench, a comprehensive benchmark with 132 articulated scenes and 449 task variations across four household domains. It defines five generalization levels: random placement, cross-part, cross-instance, cross-category, and long-horizon composition, enabling systematic evaluation beyond single-step actions.
    
    \item We propose ArtiBrain, a hierarchical framework that unifies high-level reasoning and low-level control. It comprises a VLM-based Task Reasoner for structured subgoal generation and a Hybrid Controller that integrates two carefully designed policies and an Affordance Memory Bank that enables continual refinement of transferable part-level priors.

    \item We show that ArtiBrain achieves strong part-level generalization on the proposed ArtiBench, outperforming 3D-LOTUS++~\cite{garcia20253dlotus} by \textbf{67\%} on novel-part manipulation. It further achieves the best performance on long-horizon articulated tasks among the evaluated baselines, demonstrating consistent generalization from short single-step interactions to complex multi-step manipulation. 

\end{itemize}

%% file: sec/2_related_work.tex
\section{Related Work}

\paragraph{Foundation Models for Task Reasoning and Planning.}
Large Language Models (LLMs) and Vision Language Models (VLMs) enable high-level reasoning, task decomposition, and open-vocabulary planning in robotics~\cite{brohan2023can, huang2022language, sharma2022skill, zawalski2024robotic, zhang2025embodied, shi2025hi}. Prior efforts ground outputs from LLMs or VLMs to executable actions via feasibility value functions~\cite{brohan2023can}, language-to-skill translation~\cite{huang2022language, sharma2022skill}, or multimodal closed-loop prompting~\cite{zawalski2024robotic, zhang2025embodied, shi2025hi}.
However, they still reason mostly at the semantic level and lack explicit knowledge on how to contact and how to move after contact, resulting in unreliable behavior in contact-rich and multi-part manipulation settings.
In this work, we bridge this gap by coupling VLM reasoning with articulation-aware control and part-aware affordance transfer mechanism, enabling high-level plans to produce physically feasible action trajectories.

\noindent\textbf{Diffusion Policies for Visuomotor Control.}
Diffusion-based visuomotor policies generate actions through conditional denoising. They have shown strong stability and support multimodal action reasoning~\cite{chi2023diffusion, ha2023scaling, team2024octo}.
Recent 3D extensions incorporate point clouds and multi-view features to enhance spatial grounding~\cite{ze2024dp3, ke2024diffuseractor}.
Affordance-conditioned variants further inject contact priors to improve contact precision~\cite{zhao2025anchordp3, wu2025afforddp}.
However, most diffusion policies are still short-horizon and lack hierarchical structure, making it difficult to generalize across articulated parts and diverse tasks.
In this work, we combine VLM-guided hierarchical planning with an articulation-aware diffusion controller to achieve scalable and transferable manipulation in complex articulated environments.

\begin{table*}[t]
\centering
\scriptsize
\caption[Comparison of Benchmarks]
{Comparison of benchmarks for vision-and-language robotic manipulation.
For \textit{Train}, \textit{Test (short)} and \textit{Test (long)}, number outside the parentheses is the count of articulation-related tasks and the numbers in parentheses denote the count of articulation-related task variations included in training and evaluation.
\textit{Multi}: multiple action primitives;
\textit{Transfer}: unseen action–object combinations;
\textit{Atc-Task}: articulation-related test tasks;
\textit{Atc-Var}: articulation-related test task variations;
 \circled{PL} / \circled{PA} / \circled{I} / \circled{C}: generalization to unseen placements/ articulated parts / instances / categories;
\circled{L}: long-horizon task compositions.
}
\label{tab:benchmark_comparison}
\resizebox{\linewidth}{!}{
\begin{tabular}{@{}l|ccccccccccccc@{}}
\toprule
\textbf{Benchmark} & \textbf{Simulator} & \textbf{Train} & \textbf{Test (short)} & \textbf{Test (long)} & \textbf{Multi} & \textbf{Transfer} & \textbf{Atc-Task} & \textbf{Atc-Var} & \textbf{Generalization} \\
\midrule
RLBench-74~\cite{james2019rlbench} & RLBench & 1 (1) & 19(23) & 1 (1) & \cmark & \xmark & 20 & 24 & \fixedcircled{PL} \\
VLMBench~\cite{zheng2022vlmbench} & RLBench & 3 (7) & 2 (4) & 0 (0) & \cmark & \xmark & 5 & 11 & \fixedcircled{PL} \fixedcircled{I} \\
Calvin~\cite{mees2022calvin} & PyBullet  & 2 (8) & 2 (2) & 2 (2) & \cmark & \xmark & 2 & 8 & \fixedcircled{L}\\
Colosseum~\cite{pumacay2024colosseum} & RLBench & 3 (5) & 3 (5) & 1 (1) & \cmark & \xmark & 4 & 6 & \fixedcircled{PL} \\
GEMBench~\cite{garcia20253dlotus} & RLBench & 9 (11) & 9 (24) & 2 (4) & \cmark & \cmark & 27 & 39 & \fixedcircled{PL} \fixedcircled{PA} \fixedcircled{I} \fixedcircled{C} \fixedcircled{L} \\
\midrule
\textbf{ArtiBench-S (Ours)} & RLBench & 9 (12) & 16 (22) & 8 (14) & \cmark & \cmark & 24 & 36 & \fixedcircled{PL} \fixedcircled{PA} \fixedcircled{I} \fixedcircled{C} \fixedcircled{L}\\
\textbf{ArtiBench (Ours)} & RLBench & \textbf{28 (31)} & \textbf{98 (300)} & \textbf{34 (149)} & \cmark & \cmark & \textbf{132} & \textbf{449} & \fixedcircled{PL} \fixedcircled{PA} \fixedcircled{I} \fixedcircled{C} \fixedcircled{L}
\\

\bottomrule
\end{tabular}
}
\vspace{-8pt}
\end{table*}

\noindent\textbf{Articulated Object Manipulation Benchmarks.}
Standardized benchmarks have driven progress in generalizable manipulation, with simulators such as \textit{RLBench}~\cite{james2019rlbench}, \textit{CALVIN}~\cite{mees2022calvin}, \textit{ManiSkill}~\cite{mu2021maniskill}, \textit{iGibson2.0}~\cite{li2021igibson}, and \textit{AI2-THOR}~\cite{kolve2022ai2thor} enabling reproducible and scalable evaluation of robotic policies. \textit{GEMBench}~\cite{garcia20253dlotus}, introduces graded generalization splits across instances and categories to assess policy robustness.
However, existing benchmarks vary significantly in terms of physics fidelity and generalization scope. \textit{RLBench}~\cite{james2019rlbench} focuses on rigid-object manipulation without articulation reasoning, while \textit{VLMBench}~\cite{zheng2022vlmbench} extends it with vision–language grounding but remains limited in scale and articulation coverage. \textit{CALVIN}~\cite{mees2022calvin} introduces long-horizon reasoning yet contains few assets and limited task diversity. \textit{Colosseum}~\cite{pumacay2024colosseum} targets robustness to environmental variations such as lighting and viewpoint changes rather than generalization across manipulable objects. Although \textit{GEMBench}~\cite{garcia20253dlotus} expands toward unseen objects and novel tasks, it still lacks systematic part-level and articulation-oriented variations.
Our ArtiBench is the first comprehensive benchmark for articulated-object manipulation, offering over 400 unique, systematic part-level and articulation-oriented variations across drawers, cabinets, appliances, and tools.

%% file: sec/3_benchmark.tex
\section{ArtiBench}
\label{sec:artibench}

Our ArtiBench systematically evaluates the generalization capability of articulated-object manipulation across short-horizon and long-horizon tasks. We consider four levels of short-horizon generalization: random placement, cross-part, cross-instance, and cross-category. These generalization primitives are further composed with rigid-object manipulation tasks to form challenging long-horizon activities. Such activities require articulation-aware interaction across multiple generalization axes, as well as reliable language-conditioned task decomposition and sequential execution. An overview is shown in Fig.~\ref{fig:artibrain-overview}.

\begin{figure}[t]
\centering
\includegraphics[width=\linewidth]{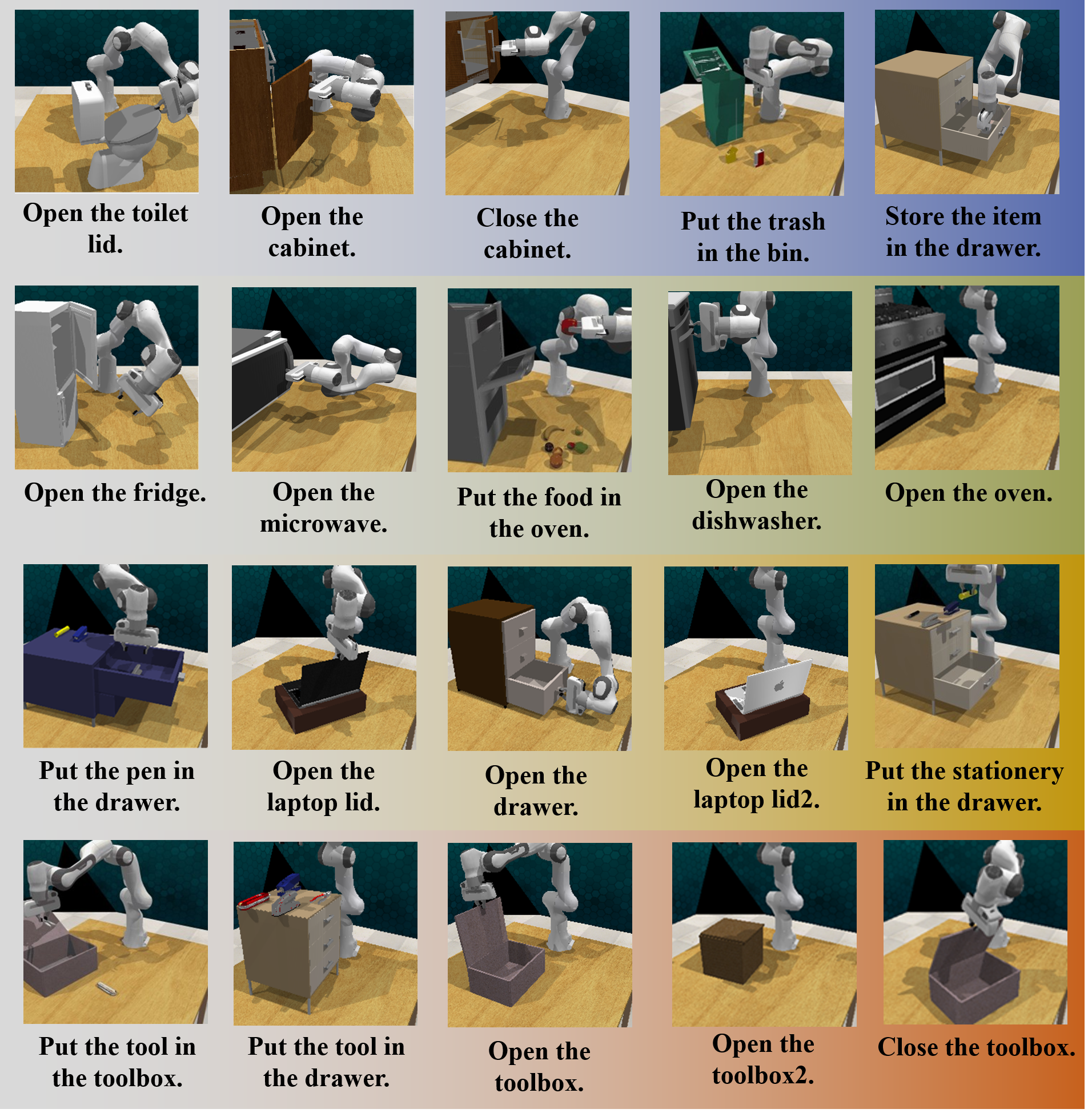}
\caption{Representative tasks from the four ArtiBench scenarios.
Examples include disposing trash and organizing items in \emph{Storage}, opening a refrigerator or oven in the \emph{Kitchen}, manipulating drawers and laptops in the \emph{Office}, and placing tools in the \emph{Tool} setting.
These tasks illustrate the diversity of everyday articulated interactions.}
\label{fig:artibench-scenarios}
\vspace{-8pt}
\end{figure}

\noindent\textbf{Benchmark Scope and Coverage.}
ArtiBench unifies articulated and rigid-object tasks across four scenarios: \emph{Kitchen}, \emph{Storage}, \emph{Office}, and \emph{Tool}. It includes diverse articulated categories such as drawers, refrigerators, ovens, toilets, and laptops, along with common storage and office items. All assets are derived from \textit{PartNet-Mobility}~\cite{xiang2020sapien}, \textit{RLBench}~\cite{james2019rlbench}, and \textit{YCB}~\cite{Calli2015YCB}, and all tasks are implemented in \textit{CoppeliaSim} via \textit{PyRep} to ensure consistent dynamics and reproducible demonstrations. Our benchmark comprises \textbf{98} short-horizon and \textbf{34} long-horizon tasks with \textbf{449} articulation-related variations, providing a comprehensive evaluation of prismatic and revolute motion patterns across realistic household and office scenarios. Its compact subset, ArtiBench-S, offers a standardized lightweight version that maintains the full evaluation protocol while focusing on the most common articulated-object operations. Representative tasks from the four scenarios are illustrated in Fig.~\ref{fig:artibench-scenarios}.

\noindent\textbf{Generalization Axes.} 
Tasks are structured along five axes that characterize progressive generalization challenges.

\textbf{(L0) Novel Placements} \circled{PL}: Tasks involve the same objects and actions as in training but under new spatial configurations, varying initial poses, positions, or distractor arrangements to assess robustness to scene perturbations.

\textbf{(L1) Novel Parts} \circled{PA}: The robot must manipulate different articulated parts of a known instance, such as opening the upper, middle, or lower drawer of the same cabinet, thereby evaluating part-level affordance transfer.

\textbf{(L2) Novel Instances} \circled{I}: This level tests generalization across unseen instances within the same category that differ in geometry, scale, or joint limits, for example adapting from a three-drawer cabinet to one with four drawers.

\textbf{(L3) Novel Categories \circled{C}}: Tasks require transferring learned manipulation strategies to entirely new articulated categories, for instance generalizing skills learned from laptops to ovens or grills.

\textbf{(L4) Novel Long-Horizon Tasks} \circled{L}: 
The most challenging level involves composing multiple short-horizon primitives from previous levels into temporally extended task sequences. 
Typical examples include multi-step procedures such as \emph{open\_drawer} $\rightarrow$ \emph{pick/place item}, or \emph{open\_oven} $\rightarrow$ \emph{insert/remove tray}. 
Short-horizon tasks (L0--L3) in ArtiBench-S are summarized in Tab.~\ref{tab:short_horizon}. Representative long-horizon tasks (L4) from the full ArtiBench are presented in Tab.~\ref{tab:long_tasks}.
Comprehensive task lists and configuration details are provided in the supplementary material.
\begin{table}[t]
\centering
\caption{Short-horizon task coverage (L0--L3) in ArtiBench-S.
}
\label{tab:short_horizon}
\resizebox{\linewidth}{!}{
\begin{tabular}{@{}l|l l l l@{}}
\toprule
\textbf{Primitive} & \textbf{L0} & \textbf{L1} & \textbf{L2} & \textbf{L3} \\
\midrule
\textbf{Open} &
\begin{tabular}[c]{@{}l@{}}Drawer top\\ Drawer bottom\\ Drawer2 top\\ Drawer2 bottom\end{tabular} &
\begin{tabular}[c]{@{}l@{}}Drawer middle\\ Drawer2 middle\end{tabular} &
\begin{tabular}[c]{@{}l@{}}DrawerSmall\\ DrawerLong\end{tabular} &
-- \\
\midrule
\textbf{Open (revolute)} &
\begin{tabular}[c]{@{}l@{}}Oven3\\ Box\end{tabular} &
\begin{tabular}[c]{@{}l@{}}--\end{tabular} &
-- &
\begin{tabular}[c]{@{}l@{}}Trashcan\\ Toilet\end{tabular} \\
\midrule
\textbf{Close (prismatic)} &
\begin{tabular}[c]{@{}l@{}}Drawer top\\ Drawer bottom\end{tabular} &
\begin{tabular}[c]{@{}l@{}}Drawer middle\end{tabular} &
-- & -- \\
\midrule
\textbf{Close (revolute)} &
\begin{tabular}[c]{@{}l@{}}Laptop Lid\\Microwave\\ Fridge\\ Cabinet2\end{tabular} &
-- &
\begin{tabular}[c]{@{}l@{}}Cabinet5\\ Microwave2\end{tabular} &
\begin{tabular}[c]{@{}l@{}}Toilet\end{tabular} \\
\bottomrule
\end{tabular}
}
\vspace{-8pt}
\end{table}

\begin{table}[t]
\centering
\caption{Representative long-horizon tasks (L4).
Each task composes sequential primitives involving both articulated and rigid-object interactions.
Comprehensive task configurations and variations are detailed in the supplementary material.}
\label{tab:long_tasks}
\resizebox{\linewidth}{!}{
\begin{tabular}{@{}c|l|l|c@{}}
\toprule
\textbf{Domain} & \textbf{Task} & \textbf{Sub-task chain} & \textbf{Var.} \\
\midrule
Storage & bin\_dispose\_trash & open\_bin $\to$ pick\_trash $\to$ place\_in\_bin & 3 \\
Kitchen & oven\_place\_food & open\_oven $\to$ pick\_food $\to$ place\_in\_oven & 8 \\ 
Office & drawer\_store\_stationery & open\_drawer $\to$ pick\_stationery $\to$ place\_in\_drawer & 12 \\ 
Tools & toolbox\_store\_tools & open\_toolbox $\to$ pick\_tool $\to$ place\_in\_box & 1 \\ 
\bottomrule
\end{tabular}}
\vspace{-13pt}
\end{table}

%% file: sec/4_method.tex
\section{ArtiBrain}
\label{sec:method}
While existing visuomotor policies excel in short-horizon or single-object settings, they struggle to reason over sequential dependencies and to coordinate different control regimes required by rigid and articulated objects. 
Different from prior modular or monolithic frameworks~\cite{shridhar2022cliport, zitkovich2023rt2, duan2024manipulate, zhao2025anchordp3}, 
our ArtiBrain has a novel hierarchical architecture, that unifies open-vocabulary reasoning, hybrid control, and adaptive affordance transfer within a closed-loop pipeline.

At each timestep~$t$, our agent receives multi-view RGB-D images~$I_t$, point clouds~$P_t$, and proprioceptive states~$S_t$. 
The objective is to learn a policy $\pi(a_t \mid O_t, l)$, where $O_t = \{I_t, P_t, S_t\}$ and $l$ denotes an open-vocabulary instruction. 
The action $a_t$ specifies the robot control command, including end-effector motion and gripper actuation. 
Unlike previous end-to-end visuomotor models that directly regress actions from pixels, 
ArtiBrain explicitly decomposes decision-making into reasoning, control, and memory components, 
enabling interpretable, generalizable, and self-correcting behavior in long-horizon articulated manipulation.

Our ArtiBrain consists of three modules:
\textbf{(i) VLM-based Task Reasoner.} A vision–language model serves as an embodied semantic planner that parses open-vocabulary instructions into structured sub-tasks with success conditions, predicts manipulation type, and issues real-time verification for adaptive replanning or retrying, enabling grounded and interpretable task reasoning.
\textbf{(ii) Hybrid Controller.} A novel dual-branch controller dynamically selects between a geometry-guided keyframe policy (GeoKeyframe) for structured rigid-object motions and an affordance-guided diffusion policy (ArtiDiffusion) for contact-rich articulated interactions, achieving both efficiency and generalization.
\textbf{(iii) Affordance Memory Bank.} A self-expanding memory accumulates verified part-level affordances from successful interactions, enhancing transfer across parts and categories.

\subsection{VLM-based Task Reasoner}
\label{sec:vlm}
Unlike conventional perception modules that passively interpret visual input, the VLM in our ArtiBrain is designed as an \textit{embodied semantic reasoning agent} that bridges natural-language intent and executable robot control.
Before execution, it analyzes the current scene to decompose the instruction into semantic sub-tasks and select the most informative observation view from multi-view inputs to guide downstream controllers.
Given the natural-language instruction $l$ and initial observation $I_0$, the VLM produces a structured plan $\Pi$:
\begin{equation}
\Pi = [(p_i, o_i, c_i)]_{i=1}^N,
\label{eq:plan}
\end{equation}
where $\Pi$ denotes an ordered sequence of $N$ sub-tasks,
$p_i$ represents a primitive action type, such as \textit{open} or \textit{pick},
$o_i$ specifies the semantically grounded target object,
and $c_i$ denotes the corresponding success condition.

\begin{figure}[t]
    \centering
    \includegraphics[width=0.95\linewidth]{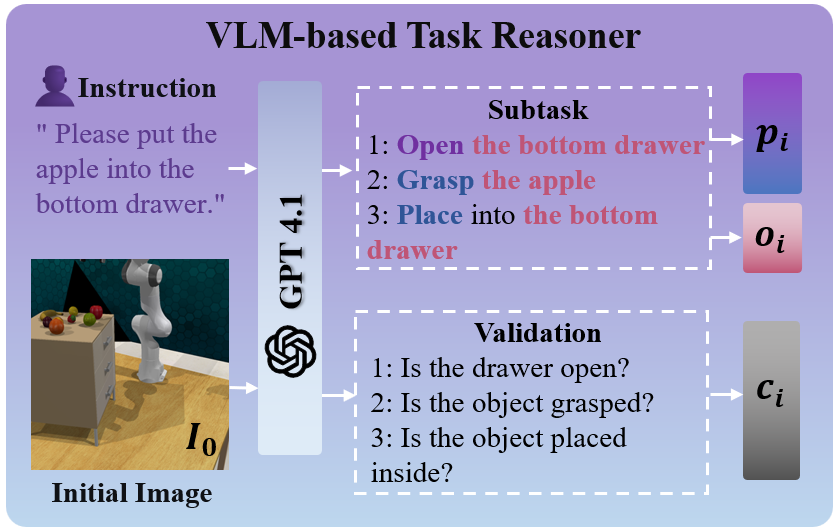}
    \caption{Architecture of our VLM-based Task Reasoner in ArtiBrain.  
    Given a natural-language instruction and initial observation $I_0$, the VLM generates a structured plan of sub-tasks $(p_i, o_i)$ with corresponding success conditions $c_i$. The reasoning process ensures each action is executed and validated before progressing.}
    \label{fig:task_reasoner}
    \vspace{-8pt}
\end{figure}

An overview of this reasoning and execution process is illustrated in Fig.~\ref{fig:task_reasoner}. As shown, the VLM decomposes the instruction into actionable sub-tasks, while also generating the corresponding success validation criteria for each task. This structured reasoning enables dynamic adaptation to task complexities and ensures that each action is performed correctly before moving to the next step.

During execution, our controller can select the motion branch for each primitive $p_i$ and performs the action. After each step, the verifier checks whether the success condition $c_i$ is met using visual feedback and task-specific cues such as grasp detection, drawer displacement, or object placement. If $c_i$ is unsatisfied, the controller refines the motion; once satisfied, the system advances to the next subtask, maintaining closed-loop, robust progression. Verified articulated interactions are stored in the Affordance Memory Bank, enabling continual improvement and long-horizon execution.

\subsection{Hybrid Controller}
\label{sec:control}

\begin{figure*}[t]
\centering
\captionsetup{type=figure}
\includegraphics[width=\textwidth]{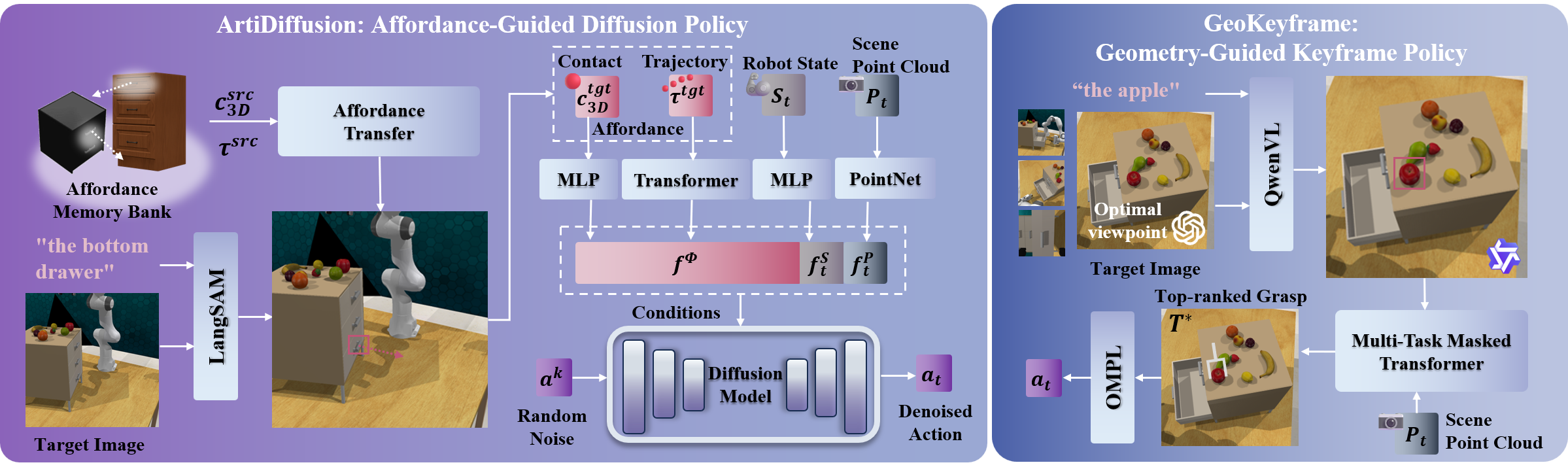}
\vspace{-20pt}
\captionof{figure}{
Architecture of the Hybrid Controller in our ArtiBrain. 
The controller integrates two branches: 
ArtiDiffusion for articulated object manipulation, employing a four-encoder architecture to extract point cloud features from $P_t$, encode robot state $S_t$, and process transferred affordance $\Phi^{\mathrm{tgt}} = (c^{\mathrm{tgt}}_{\mathrm{3D}}, \tau^{\mathrm{tgt}})$ obtained by geometrically aligning retrieved source affordance $\Phi^{\mathrm{src}} = (c^{\mathrm{src}}_{\mathrm{3D}}, \tau^{\mathrm{src}})$. The fused features condition a diffusion policy that generates action $\mathbf{a}_t$ through temporal U-Net denoising of noised sequence $\mathbf{a}^k$; GeoKeyframe for rigid objects, selecting optimal grasp pose $\mathbf{T}^\star$ and generating action $\mathbf{a}_t$ via geometric planning.
}
\label{fig:artibrain-hybrid-controller}
\vspace{-6pt}
\end{figure*}

To execute each primitive $p_i$, we employ a novel hybrid controller that selects the motion strategy based on the manipulation mode inferred by the VLM. We formalize the control policy as:
\begin{equation}
a_t = \mathcal{C}_{p_i}(o_i, I_t, D_t, S_t),
\qquad
\mathcal{C}_{p_i} \in \{\mathcal{C}_{\text{rigid}},\, \mathcal{C}_{\text{art}}\},
\label{eq:hybrid-control}
\end{equation}
where $p_i$ denotes the VLM-predicted action primitive. $o_i$ is the VLM-inferred target object for the current scene and action primitive. $I_t$ and $D_t$ represent the current RGB-D observations from the camera, from which the corresponding point cloud $P_t$ can be derived. $S_t$ denotes the current robot state. An overview of the Hybrid Controller is presented in Fig.~\ref{fig:artibrain-hybrid-controller}.

For rigid-object manipulation, our keyframe-based controller $\mathcal{C}_{\text{rigid}}$ (GeoKeyframe) leverages kinematic priors for efficient trajectory planning. For articulated manipulation, we propose a novel part-aware affordance-guided diffusion controller $\mathcal{C}_{\text{art}}$ (ArtiDiffusion) that retrieves from the affordance memory to generate continuous actions for contact-rich motion planning.
These two controllers are dynamically selected by the VLM, enabling robust execution across diverse manipulation scenarios.
\subsubsection{GeoKeyframe: Rigid Object Branch}
For rigid manipulation actions, we adopt a perception--action pipeline that couples open-vocabulary grounding with 6-DoF grasp synthesis. GPT-4.1 selects the optimal viewpoint from four camera views, and Qwen-VL~\cite{Qwen-VL} localizes the target object in the selected image. Multi-Task Masked Transformer (M2T2)~\cite{yuan2023m2t2} operates on the full-scene point cloud and predicts grasp hypotheses across the workspace, each parameterized by an end-effector pose and confidence score. To isolate grasps associated with the grounded target, we back-project the 2D bounding box into 3D and perform KD-Tree nearest-neighbor filtering to retain only grasps whose contact points lie within the target region. Remaining candidates are further pruned through 3D collision checks, yielding a feasible set $\mathcal{K}$. The final grasp is selected as the highest-confidence feasible prediction:
\vspace{-3pt}
\begin{equation}
\mathbf{T}^\star = \arg\max_{\mathbf{T}_k \in \mathcal{K}} s_k,
\label{eq:best-grasp}
\end{equation}
where $s_k$ denotes the confidence score predicted by the grasp network for candidate $\mathbf{T}_k$. The collision-free trajectory to $\mathbf{T}^\star$ is then computed and executed by a motion planner using the Open Motion Planning Library (OMPL)~\cite{Sucan2012OMPL}, enabling zero-shot grasping of unseen rigid objects.

\subsubsection{ArtiDiffusion: Articulated Object Branch}

Articulated object manipulation presents a core challenge in robotic control, as it requires accurate contact reasoning and reliable motion planning. To tackle this, we introduce a conditional diffusion policy grounded in retrieved affordance memory, enabling dynamic and contact-aware motion synthesis.

\noindent\textbf{Architecture.}
The policy comprises four encoders and a U-Net-based denoising network. 
We extract point cloud features $f^P_t \in \mathbb{R}^{d_P}$ via PointNet~\cite{qi2016pointnet}, 
encode robot state $S_t$ into $f^S_t \in \mathbb{R}^{d_S}$ via an MLP, and 
process affordances $\Phi = (c, \tau)$ through two encoders: the contact point 
$c$ is encoded into $f^\Phi_c \in \mathbb{R}^{d_c}$ via an MLP, while the 
trajectory sequence $\tau$ is encoded into $f^\Phi_\tau \in \mathbb{R}^{d_\tau}$ 
via a Transformer encoder. These features are concatenated as the global 
conditioning vector $\mathbf{f}_t = [f^P_t, f^S_t, f^\Phi_c, f^\Phi_\tau]$.

The denoising network $\epsilon_\theta$ is a 1D temporal U-Net that processes 
the initial noised action sequence $\mathbf{a}^k_t \in \mathbb{R}^{H \times d_a}$. Global conditions 
$\mathbf{f}_t$ are injected at each layer via Feature-wise Linear Modulation (FiLM)~\cite{perez2018film}, 
which adaptively modulates the intermediate features based on the concatenated 
context. Combined with sinusoidal embeddings for the diffusion timestep $k$, 
the network outputs predicted noise $\epsilon_\theta(\mathbf{a}^k_t, k, \mathbf{f}_t)$.

\noindent\textbf{Training and Inference.}
During training, for each sample in the batch, we randomly sample a diffusion 
timestep $k \sim \text{Uniform}(1,K)$ and Gaussian noise $\epsilon^k \sim \mathcal{N}(0, I)$. 
We then minimize the mean squared error (MSE) between the actual noise and the 
network's prediction:
\begin{equation}
\mathcal{L} = \|\epsilon^k - \epsilon_\theta(\mathbf{a}_{t}^{\,k}, k, \mathbf{f}_t)\|^2,
\label{eq:diffusion-loss}
\end{equation}
where 
$\mathbf{a}_{t}^{\,k}
= \sqrt{\alpha_k}\,\mathbf{a}_{0}
+ \sqrt{1-\alpha_k}\,\epsilon^k$,
is the noised action formed by adding $\epsilon^k$ to ground truth $\mathbf{a}_0$, and 
the loss is computed as the MSE averaged over the training batch.

At inference, we apply DDIM~\cite{song2021denoising} sampling for efficient denoising from 
$\mathbf{a}^{K}_{t} \sim \mathcal{N}(0, I)$:
\begin{equation}
\mathbf{a}^{k-1}_{t} 
= 
\sqrt{\alpha_{k-1}}
\left(
\frac{
    \mathbf{a}^{k}_{t} - \sqrt{1-\alpha_{k}}\,\hat{\epsilon}^{\,k}_{t}
}{
    \sqrt{\alpha_{k}}
}
\right)
+
\sqrt{1-\alpha_{k-1}}\,\hat{\epsilon}^{\,k}_{t},
\end{equation}
where
\(
\hat{\epsilon}^{\,k}_{t}
= 
\epsilon_\theta(\mathbf{a}^{k}_{t},\, k,\, \mathbf{f}_{t})
\).
This deterministic sampling provides fast and stable action generation under affordance-guided conditioning.

\noindent\textbf{Affordance Memory Bank.}
Inspired by AffordDP~\cite{wu2025afforddp}, We maintain a structured memory that stores successful manipulation experiences:
\begin{equation}
\mathcal{M}_\kappa = \big\{(\mathcal{T}_i, v_i, o_i, \Phi_i, I_{0,i}, P_{0,i})\big\}_{i=1}^{N_\kappa},
\label{eq:memory}
\end{equation}
where each entry includes the task $\mathcal{T}_i$, variation identifier $v_i$, target object $o_i$, affordance $\Phi_i = (c_i, \tau_i)$ with contact point $c_i$ and trajectory $\tau_i$, initial RGB crop $I_{0,i}$, and corresponding point cloud $P_{0,i}$.
The memory is divided into buckets $\mathcal{M}_\kappa$ indexed by articulation type 
$\kappa \in \{\text{revolute}, \text{prismatic}\}$.
The bucket size $N_\kappa$ denotes the number of stored demonstrations in $\mathcal{M}_\kappa$.

At test time, given a query crop $I^{\text{q}}$, we compute its embedding 
$z^{\text{q}} = \text{CLIP}(I^{\text{q}})$ and retrieve the most similar demonstration:
\begin{equation}
(\mathcal{T}^\star, v^\star, o^\star, \Phi^\star, I_0^\star, P_0^\star) = 
\arg\max_{i \in \mathcal{M}_\kappa}
\cos\!\big(z^{\mathrm{q}}, z_i^{0}\big),
\end{equation}
where cosine similarity measures visual similarity between query $z^{\text{q}}$ and stored appearances $z_i^{0}$.

Unlike the static memory system from AffordDP~\cite{wu2025afforddp}, our memory supports dynamic updates. 
After each successful task execution, we append the newly generated demonstration 
$(\mathcal{T}_i, v_i, o_i, \Phi_i, I_{0,i}, P_{0,i})$ to $\mathcal{M}_\kappa$. 
This continual update mechanism enables online adaptation to novel objects and unseen articulation configurations, 
progressively enriching the Affordance Memory Bank and improving generalization over time.

\noindent\textbf{Part-Aware Affordance Transfer.}
While existing methods like AffordDP~\cite{wu2025afforddp} transfer affordances at the object level, our approach advances this paradigm to the part level, enabling fine-grained localization and alignment of functional components such as handles or lids. The retrieved affordance $\Phi^{\mathrm{src}} = (c^{\mathrm{src}}_{\mathrm{3D}}, \tau^{\mathrm{src}})$ is transferred to the target scenario through geometric alignment guided by part-level segmentation.
Specifically, we first project the source contact point $c^{\mathrm{src}}_{3\mathrm{D}}$ onto the image plane. 
Given the target RGB-D view $(I_{\mathrm{tgt}}, D_{\mathrm{tgt}})$, 
we use LangSAM~\cite{langsam} to segment fine-grained manipulation parts, 
crop and upsample both source and target masks, 
and establish pixel-level correspondence via SD-DINOv2~\cite{zhang2024sddinov2} feature matching:
\begin{equation}
c^{\mathrm{tgt}}_{2\mathrm{D}}
=\arg\min_{\mathbf{x}^{\mathrm{tgt}}_{2\mathrm{D}}\in\Omega_{\mathrm{part}}}
\left\|
\hat f\!\left(c^{\mathrm{src}}_{2\mathrm{D}}\right)
-\hat f\!\left(\mathbf{x}^{\mathrm{tgt}}_{2\mathrm{D}}\right)
\right\|_2,
\label{eq:feature-matching}
\end{equation}
where $\Omega_{\mathrm{part}}$ denotes the pixel set within the segmented target part, 
and $\hat f(\cdot)$ is the L2-normalized SD-DINOv2~\cite{zhang2024sddinov2} feature embedding. 
The matched pixel is then back-projected with depth $D_{\mathrm{tgt}}$ to obtain the 3D contact $c^{\mathrm{tgt}}_{3\mathrm{D}}$.

For post-contact motion, we use PointSAM~\cite{zhou2024pointsam} to extract part-level point clouds 
$\mathcal{P}_{\mathrm{src}}$ and $\mathcal{P}_{\mathrm{tgt}}$ around the contact region, 
normalize them to local frames, and estimate the rigid transformation 
$T(\mathbf{x}) = R\mathbf{x} + \mathbf{t}$ 
via RANSAC-initialized ICP~\cite{fischler1981random, chen1992icp}:
\vspace{-0.6em}
\begin{equation}
\min_{R \in SO(3),\, \mathbf{t}} 
\sum_{i=1}^N 
\left\|
\mathbf{q}_i - (R \mathbf{p}_i + \mathbf{t})
\right\|_2^2,
\label{eq:registration}
\end{equation}
where $\mathbf{p}_i \in \mathcal{P}_{\mathrm{src}}$ and 
$\mathbf{q}_i \in \mathcal{P}_{\mathrm{tgt}}$ 
denote the corresponding 3D points in the source and target parts, respectively.

The trajectory is transformed accordingly, then adjusted by the local contact displacement 
$\delta = c^{\mathrm{tgt}}_{\mathrm{local}} - T(c^{\mathrm{src}}_{\mathrm{local}})$, 
and denormalized to obtain $\tau^{\mathrm{tgt}}$.

By incorporating LangSAM~\cite{langsam}-based part segmentation and local geometric registration, 
our method transfers affordance knowledge at the part level. 
Our design enables the robot to generalize manipulation strategies across distinct parts 
and objects with shared articulation semantics, 
surpassing the generalization capability of object-level frameworks such as 
AffordDP~\cite{wu2025afforddp}.

%% file: sec/5_experiments.tex
\section{Experiments}
\label{sec:experiments}
We evaluate our ArtiBrain across two settings:
(i) simulation experiments on our ArtiBench, covering progressively harder generalization over \emph{placements}, \emph{parts}, \emph{instances}, and \emph{categories}, as well as \emph{long-horizon} compositional tasks; and
(ii) Real-world experiments on a Franka Research 3 (FR3) robot, validating physical transfer and robustness in real environments. All experiments are conducted on a single NVIDIA A40 GPU (48\,GB) and the primary evaluation metric is Success Rate (SR, $\uparrow$), defined as the percentage of tasks successfully completed.

\subsection{Simulation Results}
\label{subsec:setup}

\noindent\textbf{Baselines.}
We compare ArtiBrain with three representative state-of-the-art generalization methods: 3D-LOTUS~\cite{garcia20253dlotus}, 3D-LOTUS++~\cite{garcia20253dlotus}, and AffordDP~\cite{wu2025afforddp}. 3D-LOTUS and 3D-LOTUS++~\cite{garcia20253dlotus} show strong generalization on GEMBench~\cite{garcia20253dlotus}.
AffordDP~\cite{wu2025afforddp} generalizes across instances and categories through diffusion.
All baselines use official implementations when available and are trained on ArtiBench under identical splits and evaluation settings.

\noindent\textbf{Evaluation Protocol.}
Each task in ArtiBench is trained with 50 demonstrations and evaluated across the L0–L4 generalization levels, with results averaged over three random seeds. 

\label{subsec:sim_exp}

\begin{figure}
    \centering
    \includegraphics[width=\linewidth]{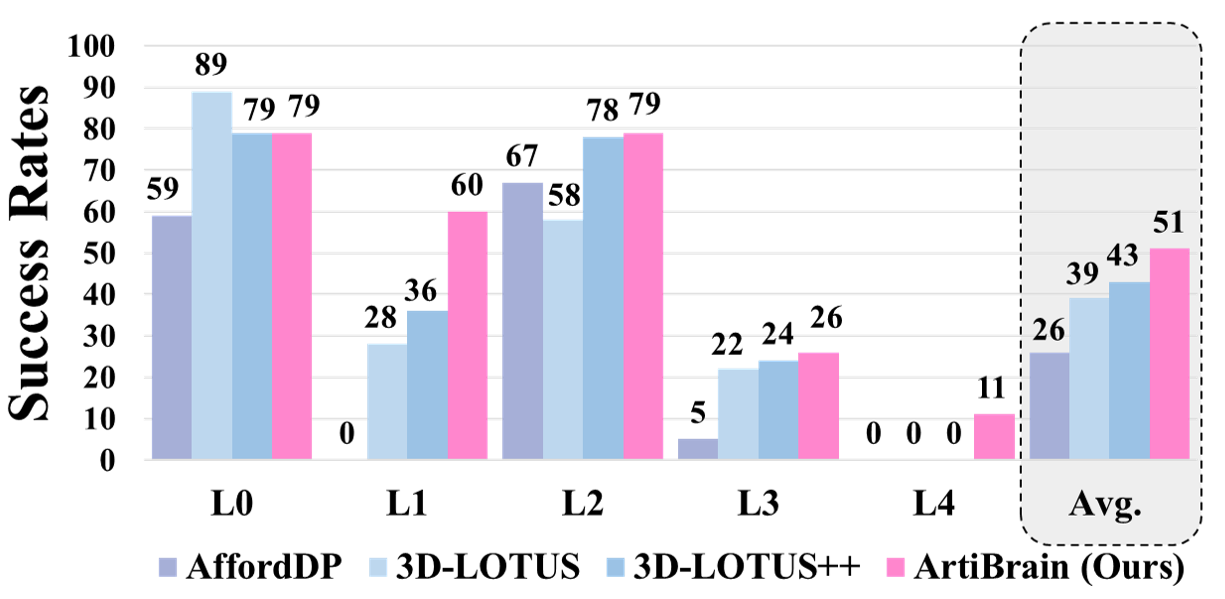}
    \caption{Results on ArtiBench.
All numbers denote success rates (\%), averaged over three random seeds.
ArtiBrain achieves the best generalization performance across L1–L4 levels.
}
    \label{fig:artibench_results}
    \vspace{-10pt}
\end{figure}
\begin{figure*}[t]
    \centering
    \begin{subfigure}[t]{0.45\textwidth}
        \centering
        \includegraphics[width=0.97\linewidth]{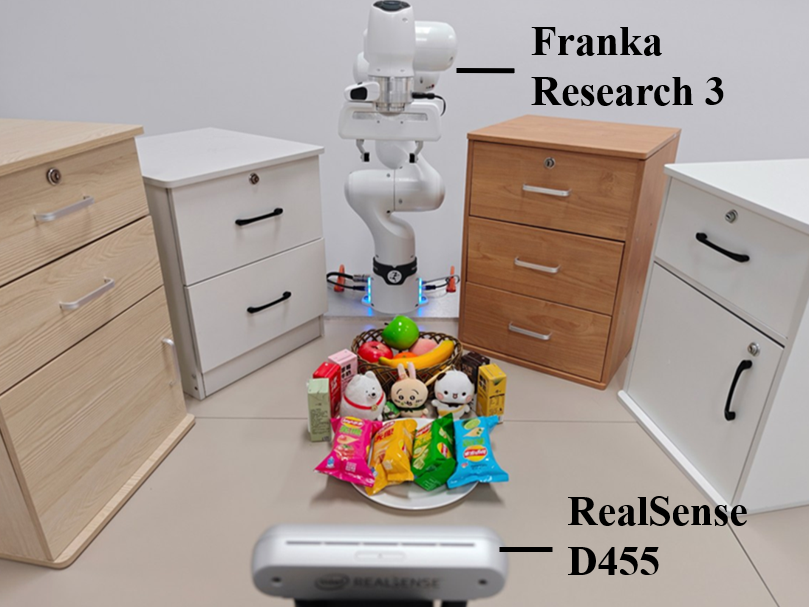}
        \caption{Real-world experimental setup.}
        \label{fig:real_a}
    \end{subfigure}
    \hspace{5pt}
    \begin{subfigure}[t]{0.45\textwidth}
        \centering
        \includegraphics[width=0.97\linewidth]{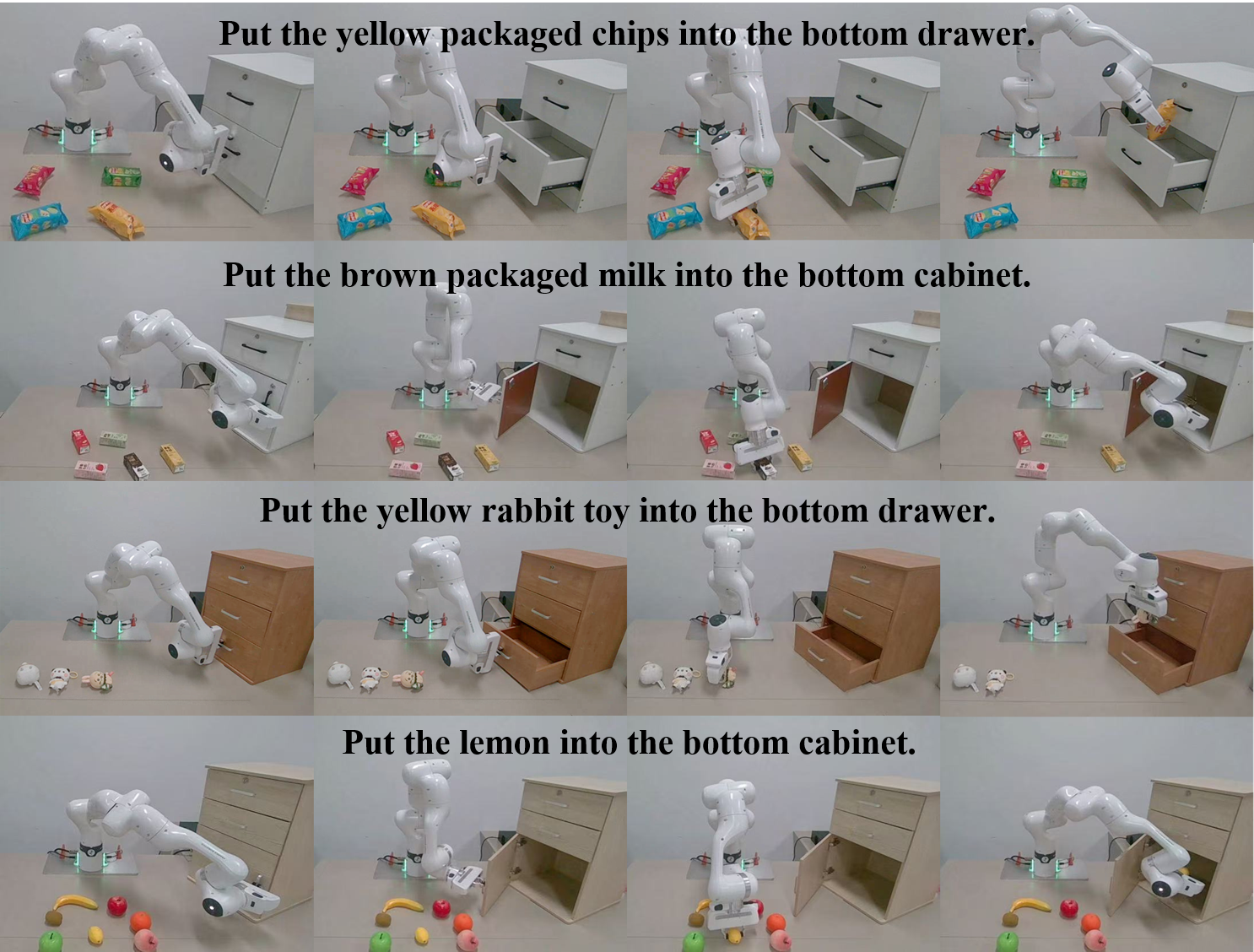}
        \caption{Long-horizon task performance.}
        \label{fig:real_b}
    \end{subfigure}

    \caption{Real-world setup and long-horizon task performance.
    (a) Experimental setup with the FR3 robot and D455 RGB-D camera. 
    (b) Example long-horizon scenarios combining articulated and rigid interactions, 
    such as placing household items into drawers and cabinets.}
    \label{fig:real_world_eval}
\vspace{-10pt}
\end{figure*}

\noindent\textbf{Results on ArtiBench.}
As shown in Fig.~\ref{fig:artibench_results}, our ArtiBrain delivers strong performance across all evaluation levels. It achieves a 60.0\% success rate on novel-part tasks (L1), outperforming the strongest baseline by 67\%. Moreover, it achieves the strongest performance on long-horizon compositional tasks (L4) and is the only method that successfully completes these tasks. ArtiBrain also achieves strong instance-level (L2) and category-level (L3) generalization, outperforming 3D-LOTUS and 3D-LOTUS++~\cite{garcia20253dlotus} on both levels and contributing to the highest overall average success rate among all compared methods. These results indicate that our part-aware affordance transfer mechanism enables robust performance under significant geometric, articulation, and temporal variation. Additionally, our ArtiBrain can directly transfer manipulation knowledge learned from short-horizon skills to long-horizon articulated tasks without training needed.

While 3D-LOTUS and 3D-LOTUS++~\cite{garcia20253dlotus} perform well on near-distribution tasks, their language-conditioned point-cloud policies are trained via behavior cloning on low-level gripper actions and therefore lack explicit mechanisms for articulated part selection or geometry-aware adaptation. Consequently, their performance drops markedly on L1–L3 when object geometry or articulation varies. Although 3D-LOTUS++~\cite{garcia20253dlotus} incorporates LLM-based planning and VLM-based grounding, its controller remains step-wise and open-loop, limiting its ability to maintain temporal consistency in multi-step articulated interactions. In contrast, ArtiBrain integrates part-aware affordance transfer with structured, closed-loop multi-step execution, achieving substantially stronger performance across L1–L4.

\subsection{Real-World Results}
\label{subsec:real}

We evaluate ArtiBrain in real-world articulated manipulation using a FR3 robot with an Intel RealSense D455 RGB-D camera providing front-view observations. The setup includes four articulated and nineteen rigid objects with both revolute and prismatic joints, as shown in Fig.~\ref{fig:real_world_eval}(a). Rigid-object pick-and-place operations are planned using the MoveIt!~\cite{coleman2014moveit} framework to reach target poses predicted by the policy. The evaluation covers two representative articulated tasks, \emph{OpenDoor} and \emph{PullDrawer}, along with long-horizon compositions that combine articulated and rigid-object interactions.

As illustrated in Fig.~\ref{fig:real_world_eval}(b), ArtiBrain performs reliably in long-horizon settings, consistently executing opening and pulling motions and completing extended tasks that involve placing household items into drawers and cabinets. These results demonstrate that our ArtiBrain transfers manipulation strategies learned from limited joint-level demonstrations to unseen articulated objects and multi-step scenarios, achieving robust generalization in real-world conditions.

\subsection{Ablation Studies}
Tab.~\ref{tab:ablation_combined} reports results on representative articulated tasks across all L1–L4 levels. Removing the VLM-based Task Reasoner causes a significant performance drop, particularly in long-horizon L4 tasks, as the model fails to decompose high-level instructions into coherent subgoals and to maintain temporal consistency. Disabling LangSAM~\cite{langsam} primarily reduces part- and instance-level generalization across L1–L3, as the model struggles to distinguish semantically similar components such as multiple drawers or cabinet doors in one scene, leading to inaccurate contact point localization. The variant without both modules performs worst across all levels, confirming that the reasoning module provides goal-directed task decomposition and temporal consistency, while LangSAM~\cite{langsam} ensures spatial grounding and affordance alignment. Their integration enables robust and geometry-aware reasoning for articulated manipulation.

\begin{table}[t]
\centering
\small
\renewcommand{\arraystretch}{1.1}
\setlength{\tabcolsep}{6pt}
\caption{Ablation study on ArtiBench.
VLM-based Task Reasoner includes task parsing, sub-goal decomposition, and success verification; 
LangSAM denotes the language-guided segmentation module.}
\label{tab:ablation_combined}
\begin{tabular}{cc|cccc}
\toprule
\textbf{\shortstack{VLM-based\\Task Reasoner}} & \textbf{LangSAM} & \textbf{L1} & \textbf{L2} & \textbf{L3} & \textbf{L4} \\
\midrule
\cmark & \cmark & \textbf{40.7} & \textbf{48.4} & \textbf{40.0}  & \textbf{12.8}\\
-- & \cmark & 28.0 & 40.8 & 40.0  & 0.0 \\
-- & --         & 0.0  & 24.3  & 11.0  & 0.0 \\
\bottomrule
\end{tabular}
\vspace{-10pt}
\end{table}

%% file: sec/7_conclusion.tex
\section{Conclusion}

In this work, we introduced ArtiBench, a structured benchmark for evaluating generalization in articulated manipulation across four scenarios and five levels of difficulty. Building on this testbed, we proposed ArtiBrain, a unified framework that couples a VLM-based Task Reasoner with a Hybrid Controller for both rigid and articulated object manipulation. Experiments on ArtiBench show that our ArtiBrain achieves state-of-the-art performance in part-level transfer and long-horizon tasks. Remaining challenges include VLM hallucination and robust affordance transfer, which we aim to address in future work.